\documentclass[a4paper,fleqn]{cas-dc}

\usepackage[numbers]{natbib}
\usepackage{booktabs}
\usepackage{vcell}
\usepackage{hyperref}
\usepackage{lipsum}

\def\tsc#1{\csdef{#1}{\textsc{\lowercase{#1}}\xspace}}
\tsc{WGM}
\tsc{QE}
\tsc{EP}
\tsc{PMS}
\tsc{BEC}
\tsc{DE}
\DeclareUnicodeCharacter{2212}{-}
\begin{document}\sloppy
\let\WriteBookmarks\relax
\def\floatpagepagefraction{1}
\def\textpagefraction{.001}
\shorttitle{Convolutional Neural Networks in Orthodontics: a review}
\shortauthors{S. Płotka et~al.}

\title [mode = title]{Convolutional Neural Networks in Orthodontics: a review}                      



\author[1]{Szymon Płotka}[type=editor,
                        auid=000,bioid=1,
                        orcid=0000-0001-9411-820X]
\fnmark[1]
\ead{plotkaszymon@gmail.com}


\address[1]{Institute of Computer Science, Warsaw University of Technology, plac Politechniki 1, 00-661, Warsaw, Poland}
\address[2]{Student Scientific Society of Maxillofacial Orthopaedics and Orthodontics, Poznan University of Medical Sciences, Bukowska 70, 60-812, Poznan, Poland}
\address[3]{Department of Maxillofacial Orthopaedics and Orthodontics, Poznan University of Medical Sciences, Bukowska 70, 60-812, Poznan, Poland}
\address[4]{Department of Oral and Maxillofacial Surgery, Medical Centre Leeuwarden, Leeuwarden, Henri Dunantweg 2, 8934 AD Leeuwarden, The Netherlands}

\address[5]{Tooploox, Teczowa 7, 53-601 Wroclaw, Poland}

\author[1]{Tomasz Włodarczyk}

\author[1]{Ryszard Szczerba}

\author[1]{Przemysław Rokita}

\author[2]{Patrycja Bartkowska}

\author[3]{Oskar Komisarek}

\author[4]{Artur Matthews-Brzozowski}

\author[1, 5]{Tomasz Trzciński}



\cortext[cor1]{Corresponding author}
\cortext[cor2]{Principal corresponding author}
\fntext[fn1]{This is the first author footnote. but is common to third
  author as well.}
\fntext[fn2]{Another author footnote, this is a very long footnote and
  it should be a really long footnote. But this footnote is not yet
  sufficiently long enough to make two lines of footnote text.}

\nonumnote{This note has no numbers. In this work we demonstrate $a_b$
  the formation Y\_1 of a new type of polariton on the interface
  between a cuprous oxide slab and a polystyrene micro-sphere placed
  on the slab.
  }

\begin{abstract}
Convolutional neural networks (CNNs) are used in many areas of computer vision, such as object tracking and recognition, security, military, and biomedical image analysis. This review presents the application of convolutional neural networks in one of the fields of dentistry - orthodontics. Advances in medical imaging technologies and methods allow CNNs to be used in orthodontics to shorten the planning time of orthodontic treatment, including an automatic search of landmarks on cephalometric X-ray images, tooth segmentation on Cone-Beam Computed Tomography (CBCT) images or digital models, and classification of defects on X-Ray panoramic images. In this work, we describe the current methods, the architectures of deep convolutional neural networks used, and their implementations, together with a comparison of the results achieved by them. 
The promising results and visualizations of the described studies show that the use of methods based on convolutional neural networks allows for the improvement of computer-based orthodontic treatment planning, both by reducing the examination time and, in many cases, by performing the analysis much more accurately than a manual orthodontist does.
\end{abstract}




\maketitle

\section{Introduction}

Convolutional neural networks (CNNs) are one type of deep learning algorithms that are used in many branches of computer vision dealing with image analysis. This type of network works very well for pattern analysis \cite{pattern} and recognition \cite{recognition}, object tracking \cite{objecttracking}, and medical image analysis \cite{mia1}, \cite{mia2}, \cite{mia3}. Recent results of medical studies using convolutional neural networks present this method as one of the future computer-aided for medical experts \cite{cnnfuture}.
Examples of such work are: mammographic image analysis \cite{Geras}, prediction of spontaneous preterm births \cite{Wlodarczyk}, \cite{Wlodarczyk2}, or the estimation of Achilles tendon healing progress \cite{Kapinski}. Orthodontics is a challenge for convolutional neural networks, where their application can help reduce the time of computer analysis through more accurate segmentation and automatic treatment planning.

Orthodontics is a branch of dentistry that deals with the treatment of malocclusion. In cooperation with maxillofacial surgery, it enables the treatment of complex dentofacial defects and severe craniofacial malformations. Orthodontic treatment is performed in patients of all ages. It consists of prophylactic treatment, the aim of which is to prevent future occlusion and functional treatment, enabling the restoration of a stable occlusion, which is a condition for the proper functioning of the stomatognathic system. Malocclusion and orofacial disorders hurt chewing, swallowing, breathing, and facial aesthetics, which affects mental health and the ability to function in society \cite{KIMSJ}.
The biggest challenge in orthodontic treatment is achieving stable results. The main causes of treatment instability are inappropriate arrests, orthopaedic instability of the mandibular condyles, and imbalance in the perioral muscles. Orthodontic treatment consists of several stages: diagnosis, treatment plan, proper treatment, and retention. The most important element of orthodontic treatment is the diagnostic process because a wrong diagnosis hurts the treatment plan.
In addition to determining the type of defect, it is also important to determine its aetiology and eliminate it if it still occurs \cite{RENSS}. A correct diagnosis requires taking the patient's medical history, performing a physical examination, examining radiological imaging such as a cephalometric image and Cone-Beam Computed Tomography (CBCT), analyzing diagnostic models, and registering them in an articulator. The entire orthodontic diagnosis process is complex and time-consuming \cite{RobertsHarry}. Its acceleration would be helpful for the orthodontist and the patient. Along with technological progress, specialized computer programs are created to improve the entire diagnostic process. However, they still require improvements and extensions to allow the practitioner to work faster \cite{BrownM}. 

In this work, we propose a review of works related to convolutional neural networks in orthodontics. Previous reviews describe the general applications of AI and Big Data to orthodontics without comparing the results and structure of networks or do not cover all imaging modalities \cite{Allareddy, Asiri, Faber, Schwendicke}. To the best of our knowledge, this is the first review of medical imaging works for planning orthodontic treatments based on convolutional neural networks and all imaging modalities used in orthodontics. 

The remainder of this work is organized in the following manner. In Sec. II, we present materials and methods used in orthodontics, such as cephalometric X-ray imaging, panoramic imaging, cone-beam computed tomography, and dental casts. In Sec. III we present the future of CNN in orthodontics. Finally, in Sec. IV we conclude the paper.

\section{Materials and methods}

This section presents work related to different types of imaging used in orthodontics. In the beginning, we present results in cephalometric X-ray examination. Next, we describe papers concerning panoramic X-ray image analysis, then cone-beam computed tomography (CBCT), and finally, works concerning dental casts models.

\subsection{Cephalometric X-ray}

\begin{figure*}[ht!, width=\textwidth]
    \centering
    \includegraphics[scale=0.75]{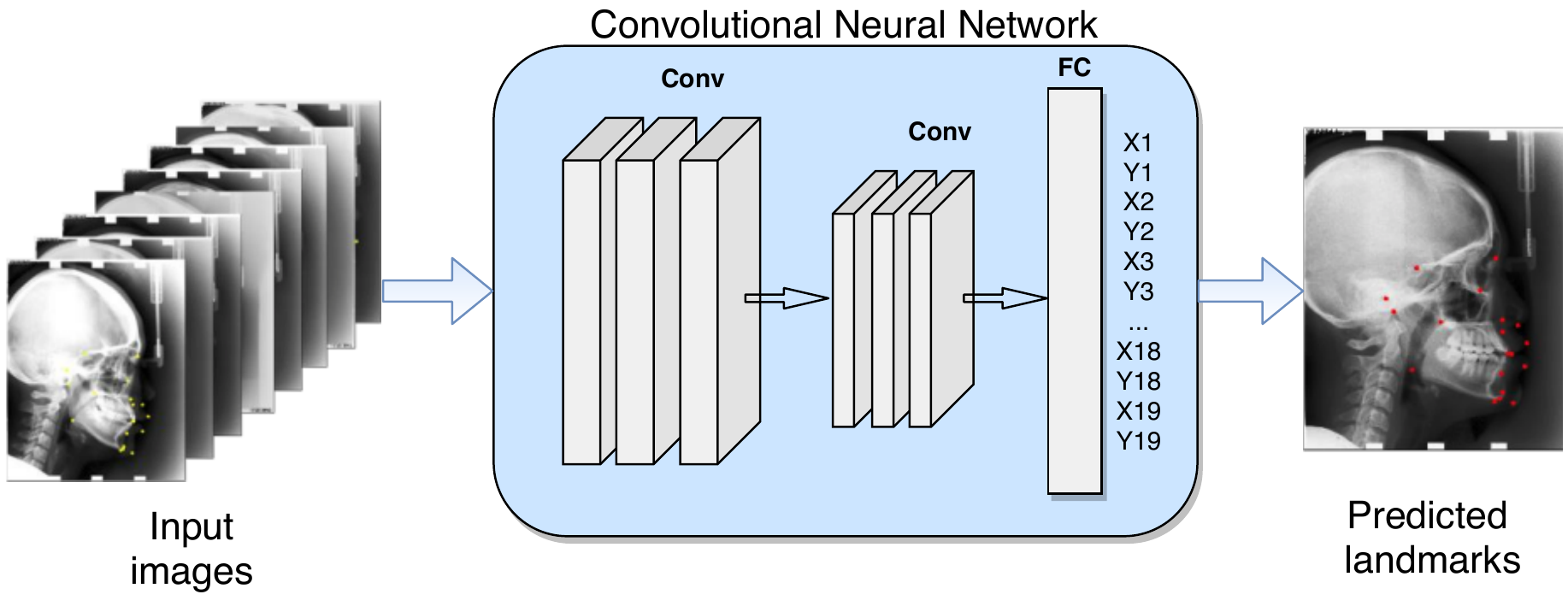}
    \caption{The sample diagram shows the steps for detecting anatomical landmarks in cephalometric X-ray images. From the left: input images as X-ray cephalometric images with annotations as anatomical points, then a convolutional neural network that processes images and annotations, in the end, predicted landmarks in the test image on the output}
    
    \label{fig:cephxdiagram}
\end{figure*}

\begin{figure}[ht!]
\begin{minipage}[b]{.49\linewidth}
  \centering
  \centerline{\includegraphics[width=4.0cm]{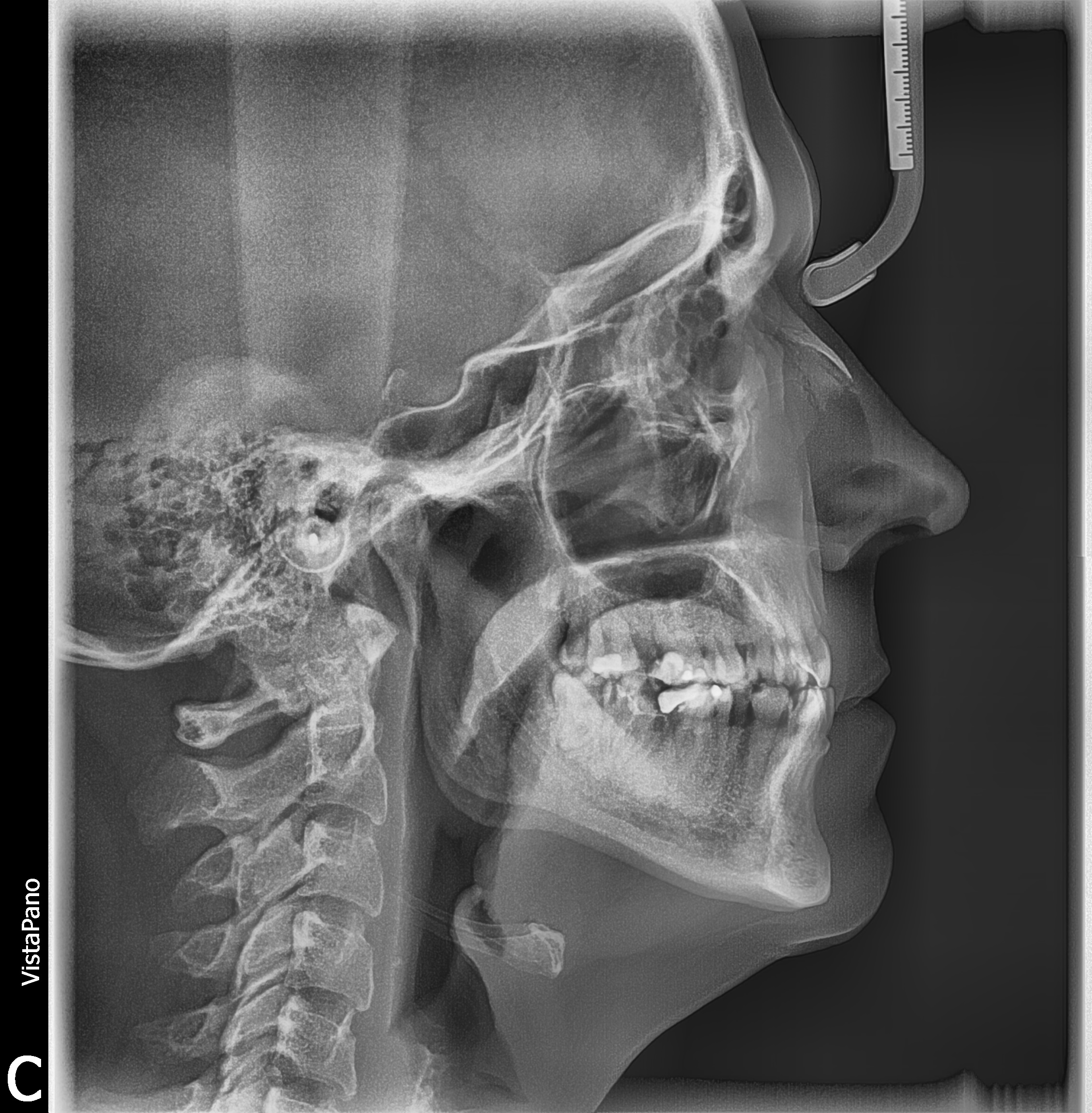}}
\end{minipage}
\hfill
\begin{minipage}[b]{.49\linewidth}
  \centering
  \centerline{\includegraphics[width=4.0cm]{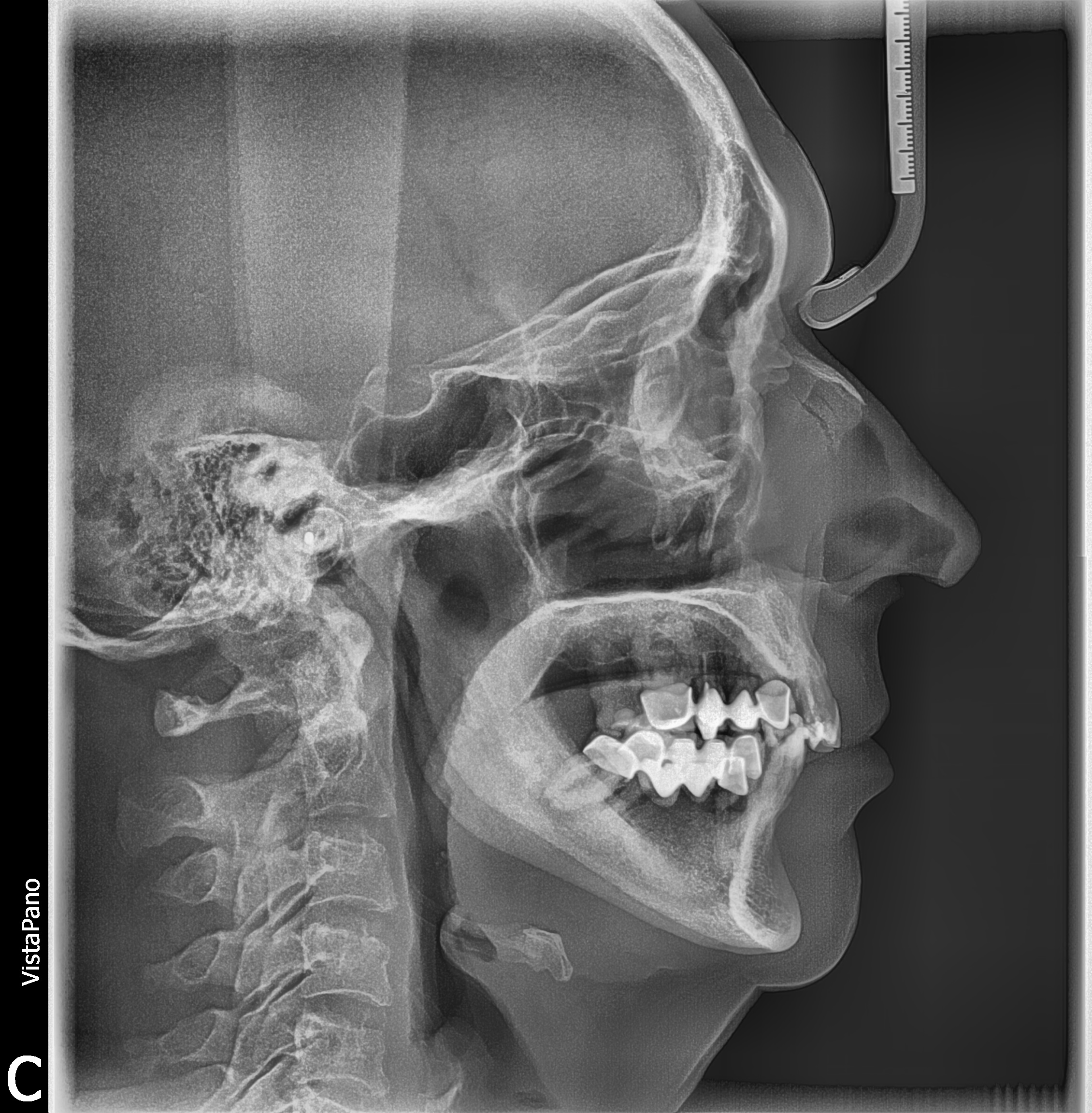}}
\end{minipage}
\caption{Example of cephalometric X-ray images}
\label{fig:cephalometric-example}
\end{figure}

Cephalometry is a type of imaging test that is an integral part of orthodontic diagnostics. It enables the correct planning of orthodontic treatment, to control its course and see the achieved results of therapy. The lateral cephalometric image of the skull shows a profile of the facial skull's bone structures and soft tissues (see Fig. \ref{fig:cephalometric-example}). It provides information on the position and size of the bone bases in relation to the skull and the inclination of the tooth axis in relation to the jaw bone base. The cephalometric analysis enables the differentiation of dentoalveolar defects from skeletal defects. Additionally, it makes it possible to predict the direction of changes in the structure of the facial skull as a result of growth processes or planned orthodontic treatment. B. Holly Broadbent \cite{Broadbent} first describes the technique of recording cephalograms. Downs \cite{Downs} publishes the first radiographic cephalometric analysis. This work highlights the clinical usefulness of cephalometry. Cecil C. Steiner \cite{Steiner} provides clinicians with information concerning skeletal balance, incisal angulation, degree of crowding, and profile characteristics obtained from cephalometry images for treatment planning. Then, computerized cephalometric systems are introduced. The main advantage of the computerized system is the speed with which the calculations could be performed, compared to previous manual measurements \cite{Taub}. In the mid-20th century, the lateral cephalogram became the standard in orthodontic treatment. However, it was soon clear that the 2D representation of a 3D object has its drawbacks. Due to the overlapping of bilateral structures, anatomical asymmetries become invisible. Moreover, the image projection causes elements farther from the film to be magnified more than elements closer to the film. Additionally, errors in the patient's head positioning distort the image obtained \cite{Hans}.

The first works related to the computer detection of landmarks date back to the 20th century \cite{20cephalometric} and the beginning of the 21st century \cite{21cephalometrica}, \cite{21cephalometricb}. The authors of these works apply image processing algorithms used in modern computer vision, such as spatial spectroscopy, mathematical morphology, and edge-based techniques like the histogram of oriented gradients (HOG). In the following years, papers using more advanced algorithms were published: pattern matching, least squares regression, or random forest. The first works that started using artificial neural networks like \cite{Cellular1}, \cite{Cellular2} , and \cite{Faghi} show promising results. The use of computer-aided analysis in modern cephalometric studies is rapidly developing. The latest research shows that the use of convolutional neural networks in computer cephalometric analysis achieves better results than when the orthodontist performs them manually \cite{Qianc}, \cite{Gilmour}, \cite{Li}. The literature mainly includes numerous studies on the automatic identification of cephalometric landmarks. The development of research on cephalometric images follows the publication of an open data set containing x-ray cephalometric images with annotations of 19 anatomical points. The data was published as part of the "Automated detection and analysis for diagnosis in cephalometric x-ray image" challenge, at the International Symposium on Biomedical Imaging (ISBI) 2014 and 2015 conference. The data included studies from 400 patients with a resolution of 1935 px $\times$ 2400 px and was annotated by two independent experts. One of the competition's evaluation metrics was mean radial error. Mean Radial Error (MRE) (see Eq. \ref{eq:mre}) is the mean distance between a landmark placed by a machine and the ground truth landmark. For each landmark, it is calculated using the following equation:
\begin{equation}
    {MRE} = \frac{\sum_{i=1}^{n}R_i}{n},
    \label{eq:mre}
\end{equation}
where $\mathrm{R}=\sqrt{\Delta x^2 + \Delta y^2}$, $n$ is the number of images \cite{isbi1}, \cite{isbi2}. 

The solution to the problem of automatic search for anatomical landmarks on X-ray cephalometric images consists of several approaches: 1) regression of points using state-of-the-art convolutional networks used for classification tasks: ResNet \cite{resnet}, AlexNet \cite{alexnet}, Inception-V3 \cite{inception}, VGG-16 \cite{vgg}, 2) encoder-decoder networks such as U-Net \cite{unet}, 3) the use of state-of-the-art convolutional neural networks as a backbone for feature extraction, and then the creation of a proprietary new module responsible for improving the accuracy and reducing the prediction error of landmarks, and 4) graph convolutional neural networks.

One of the first approaches is to use the most famous convolutional neural network architectures to regress anatomical points in the form of 38 coordinates (19 X, Y pairs). The first solution to work on the set is by Arik et al. \cite{Arik}, who propose a method to estimate the probability of the presence of landmarks in a given part of an image. This paper obtains better results in the form of greater accuracy of estimation and lower measurement error than previous approaches which use classical machine learning algorithms such as random forests \cite{randomforest}. Accuracy is the ratio of how many correct predictions the model made out of the total number of predictions. The equation for accuracy (see Eq. \ref{eq:accuracy}) is as follows:
\begin{equation}
    \frac{TP + TN}{TP+TN+FP+FN},
    \label{eq:accuracy}
\end{equation}
where $TP$, $TN$, $FP$ and $FN$ are True Positive, True Negative, False Positive and False Negative respectively. A similar approach is followed by Lee et al. \cite{LeeHansang} who uses the LightNet \cite{LightNet} and MathConvNet \cite{mathconvnet} architectures as part of their experiments. Unfortunately, too much reduction of the image resolution (from 1935 px $\times$ 2400 px to 64 px $\times$ 64 px) and the lack of data augmentation contributed to the poor accuracy of the automatic detection of anatomical landmarks. However, the work shows that the use of convolutional neural networks has promising results. Song et al. \cite{Song} apply a two-step approach: firstly, based on registration results, they extract an ROI patch centered at the coarse landmark location, secondly, a pre-trained network with the backbone of ResNet-50 is used to detect landmarks on extracted ROI patches. A ResNet uses a skip connection over two convolutional layers. This allows for easier optimization of deeper networks. Nishimoto et al. \cite{Nishimoto} in their first work use convolutional neural networks on their own data set, which is created from images found on the internet. Unfortunately, the large difference in error between the training set and the test set may indicate overfitting. Their other work \cite{Nishimoto2} uses a regression neural network with 4 convolutional layers. Unfortunately, the authors do not present any numerical results, only in the form of graphs, which makes an accurate comparison with other methods used impossible. The advantage of this work is the 4-stage learning method.

Another approach is the use of encoder-decoder convolutional neural networks based on the U-Net network. A U-Net has a downsampling path and then an upsampling path. A skip connection between the downsample and upsample steps retains spatial information of the image. One of them is the work of Goutham et al. \cite{Goutham}. The advantage of this work is the use of segmentation maps. Unfortunately, the results of this study cannot be compared because only 7 landmarks were used, hence the results are unreliable.  In the works of Zhong et al. \cite{Zhong} and Oh et al. \cite{Oh} the U-Net network is also employed but to create a method called Attention-Guided regression, which estimates heatmaps in places of sought landmarks. In \cite{Oh} anatomical context weight was used, which can be compared with graph neural networks.

Other works seek to develop new methods by adding additional modules to existing neural networks based on state-of-the-art convolutional neural networks that are used to extract features from images. Chen et al. \cite{Chenc} propose an end-to-end deep learning network that automatically and accurately detects landmarks. The architecture consists of three modules: feature extraction method based on the VGG-19 network architecture, attentive feature pyramid fusion (AFPF), and the prediction module. The proposed AFPF module significantly increases the accuracy of the detection of anatomical landmarks. This module allows working on the extracted features by different layers of the neural network with different resolutions and semantics. The predictive module uses the traditional cropping patches method to predict ground truth landmarks.  Additionally, a more efficient regression-voting solution is used, consisting of a combination of heat maps and offset maps.

\begin{table*}[width=\textwidth]
\centering
\caption{Comparison of papers based on X-ray cephalometric images}
\label{tab:cephalo_table}
\begin{tabular}{llll} 
\toprule
Author   & Architecture& Evaluation   & Year   \\ 
\midrule
\vcell{Lee et al. \cite{LeeHansang}}   & \vcell{LightNet and MatConvNet} & \vcell{They didn't show any numerical results}   & \vcell{2017}   \\[-\rowheight]
\printcelltop& \printcelltop   & \printcelltop& \printcelltop  \\

\vcell{Goutham et al. \cite{Goutham}}   & \vcell{U-Net}   & \vcell{Subset of 7 landmarks, without MRE.} & \vcell{2019}   \\[-\rowheight]
\printcelltop& \printcelltop   & \printcelltop& \printcelltop  \\

\vcell{Nishimoto et al. \cite{Nishimoto2}} & \vcell{Regression neural network with 4 CNN layers} & \vcell{\begin{tabular}[b]{@{}l@{}}Reported successful detection rates. \\ No MRE for comparison. \end{tabular}} & \vcell{2020}   \\[-\rowheight]
\printcelltop& \printcelltop   & \printcelltop& \printcelltop  \\

\vcell{Arik et al. \cite{Arik}}  & \vcell{CNN} & \vcell{\begin{tabular}[b]{@{}l@{}}Accuracy: \\ I subset: 75.58\%\\ II subset: 75.37\%\\ III subset: 67.68\% \end{tabular}}   & \vcell{2017}   \\[-\rowheight]
\printcelltop& \printcelltop   & \printcelltop& \printcelltop  \\

\vcell{Lee et al. \cite{Lee} }   & \vcell{\begin{tabular}[b]{@{}l@{}}CNN-PC (patch classification),\\ CNN-PE (point estimation) \end{tabular}} & \vcell{\begin{tabular}[b]{@{}l@{}}1.32\textasciitilde{}3.5 mm error on hard tissue and \\ 1.16\textasciitilde{}4.37 mm on soft tissue \end{tabular}} & \vcell{2020}   \\[-\rowheight]
\printcelltop& \printcelltop   & \printcelltop& \printcelltop  \\

\vcell{Nishimoto et al. \cite{Nishimoto}} & \vcell{CNN} & \vcell{\begin{tabular}[b]{@{}l@{}}MRE (training) = 0.392 mm,\\ MRE (testing) = 1.702 mm \end{tabular}}& \vcell{2019}   \\[-\rowheight]
\printcelltop& \printcelltop   & \printcelltop& \printcelltop  \\

\vcell{Oh et al. \cite{Oh}}& \vcell{U-Net}   & \vcell{\begin{tabular}[b]{@{}l@{}}MRE1 = 1.181 mm, \\ MRE2 = 1.445 mm \end{tabular}} & \vcell{2020}   \\[-\rowheight]
\printcelltop& \printcelltop   & \printcelltop& \printcelltop  \\

\vcell{Chen et al. \cite{Chenc}}  & \vcell{CNN with novel module}   & \vcell{\begin{tabular}[b]{@{}l@{}}MRE1 = 1.17 mm, \\ MRE2 = 1.48 mm \end{tabular}}   & \vcell{2019}   \\[-\rowheight]
\printcelltop& \printcelltop   & \printcelltop& \printcelltop  \\

\vcell{Qian et al. \cite{Qianc}}  & \vcell{CephaNN with multi-attention mechanism}  & \vcell{\begin{tabular}[b]{@{}l@{}}MRE1 = 1.15 mm, \\ MRE2 = 1.43 mm \end{tabular}}   & \vcell{2020}   \\[-\rowheight]
\printcelltop& \printcelltop   & \printcelltop& \printcelltop  \\

\vcell{Zhong et al. \cite{Zhong}} & \vcell{U-Net}   & \vcell{\begin{tabular}[b]{@{}l@{}}MRE1 = 1.12 mm, \\ MRE2 = 1.42 mm \end{tabular}}   & \vcell{2019}   \\[-\rowheight]
\printcelltop& \printcelltop   & \printcelltop& \printcelltop  \\

\vcell{Song et al. \cite{Song}}  & \vcell{Pre-trained ResNet50}& \vcell{\begin{tabular}[b]{@{}l@{}}MRE1 = 1.077 mm,\\ MRE2 = 1.542 mm \end{tabular}}  & \vcell{2019}   \\[-\rowheight]
\printcelltop& \printcelltop   & \printcelltop& \printcelltop  \\

\vcell{Li et al. \cite{Li}}& \vcell{Graph Convolutional Neural Network (GCNN)}   & \vcell{\begin{tabular}[b]{@{}l@{}}MRE1 = 1.04 mm, \\ MRE2 = 1.43 mm \end{tabular}}   & \vcell{2020}   \\[-\rowheight]
\printcelltop& \printcelltop   & \printcelltop& \printcelltop  \\

\vcell{Gilmour et al. \cite{Gilmour}}   & \vcell{Pre-trained ResNet34}& \vcell{\begin{tabular}[b]{@{}l@{}}MRE1 = 1.01 mm, \\ MRE2 = 1.33 mm \end{tabular}}& \vcell{2020}   \\[-\rowheight]
\printcelltop& \printcelltop   & \printcelltop& \printcelltop  \\

\bottomrule
\end{tabular}
\end{table*}

A significantly different approach is used in the work of Qian et al. \cite{Qianc}, where the first work related to the detection of anatomical landmarks is proposed based on the Faster R-CNN neural network \cite{fasterrcnn}, called CephaNet. They design a multi-task loss and adopt the multi-scale training strategy to detect small landmarks. The experiments demonstrate that the application of the two-step method can achieve satisfactory results comparable to state-of-the-art methods.
Lee et al. \cite{Lee} use a convolutional neural network for patch classification and point estimation. It uses its own dataset to search for 44 landmarks.
The current state-of-the-art results belong to Gilmour et al. \cite{Gilmour}, who obtained the smallest error on both the training and test sets. For this purpose, a pre-trained ResNet34 is used. The main advantage is the multiresolution approach to learning features across all scales. 

One of the newest approaches is the use of graph convolutional neural networks. Li et al. \cite{Li} propose a solution based on graph convolutional neural networks, arguing that it effectively exploits the structural knowledge for landmark coordinate regression. This method closes the performance gap between coordinate- and heatmap-based landmark detection methods and automatically reveals physically meaningful relationships among landmarks, leading to a task-agnostic solution for exploiting structural knowledge via step-wise graph transformations.
Fig. \ref{fig:cephxdiagram} shows the analysis of anatomical landmarks in an X-ray cephalometric image using convolutional neural networks.
Table \ref{tab:cephalo_table} shows a comparison of works that use X-ray cephalometric images.

\subsection{Panoramic X-ray}

\begin{figure*}[ht!, width=\textwidth]
    \centering
    \includegraphics[width=14.0cm]{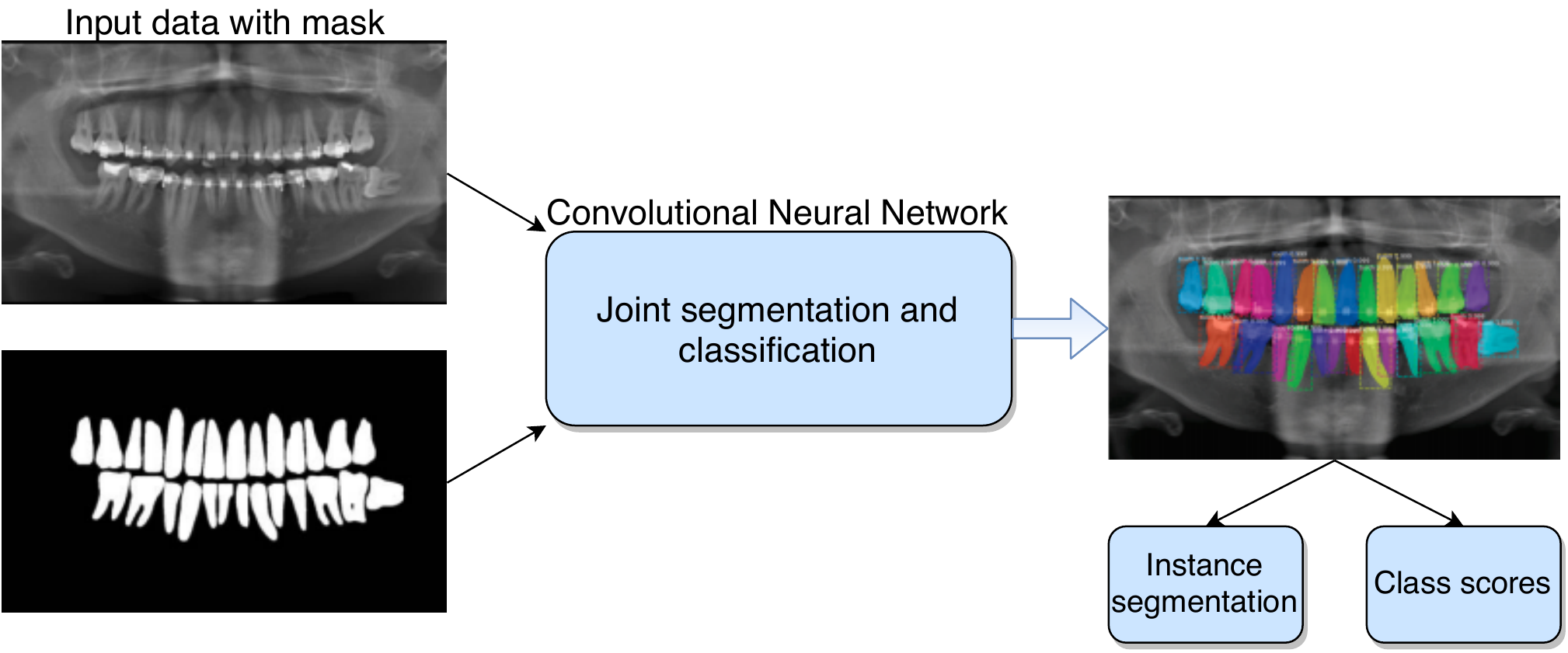}
    \caption{An example of a diagram of the joint segmentation and classification of teeth on a panoramic X-ray image. From left: panoramic X-ray image with a mask showing teeth as entry into convolutional neural network, simplified diagram of convolutional neural network and segmentation results with tooth class as output.}
    \label{fig:panoramic_xray}
\end{figure*}

\begin{figure}[ht!]
    \centering
    \includegraphics[width=7.0cm]{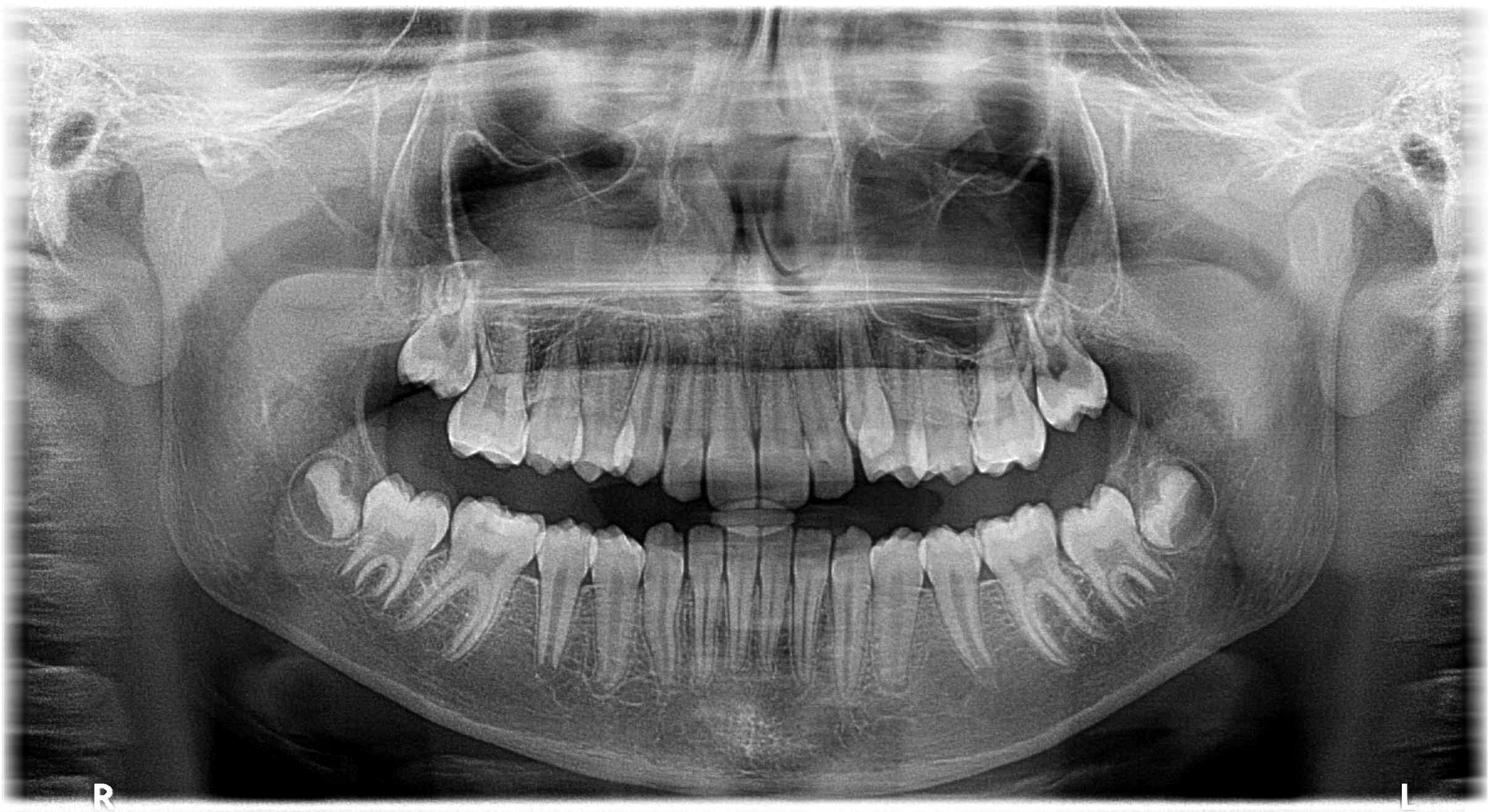}
    \caption{Example of panoramic X-ray image}
    \label{fig:pano_dataset}
\end{figure}

The first and basic image taken during orthodontic diagnostics, apart from the cephalometric image, is the pantographic image, which allows for an initial overall assessment of the stomatognathic system and finding possible pathologies. The X-ray panorama allows for the assessment of both the presence and position of tooth buds, supernumerary teeth, impacted teeth, missing tooth buds, and a comparison of both of the mandibular heads. The panoramic image is illustrative as it does not reflect the actual size of the examined structures due to uneven magnification (see Fig. \ref{fig:pano_dataset}). The panoramic technique arose from the need to image the jawbones. The narrow beam principle was described in 1922 \cite{Zulauf}. Experimental work and equipment developed in the 1950s resulted in commercially available machines in the early 1960s \cite{Hallikainen}. The paper \cite{Pano-Silva} gives a review and comparison of computer vision techniques used for the sub-task of segmentation on panoramic x-rays before CNN’s, covering methods such as region-based, threshold-based, cluster-based, boundary-based, and watershed-based. Along with the paper, a dataset of 1500 panoramic x-rays was published. It is used, with modification, in some of the following papers. From our review we exclude papers which concern forensic dentistry, to focus on the orthodontic use-case.

\begin{table*}[width=\textwidth, cols=5, pos=h]
\centering
\caption{Comparison of papers based on X-ray panoramic images}
\label{tab:panoramic_table}
\begin{tabular*}{\tblwidth}{@{} LLLLL@{} }
\toprule
Author& Architecture& \begin{tabular}[c]{@{}l@{}}Number of data\\ (Panoramic X-Ray\\ if not mentioned) \end{tabular} & \begin{tabular}[c]{@{}l@{}}Evaluation\\ (average accuracy\\ if not mentioned) \end{tabular}& Year   \\ 
\midrule
\vcell{Fukuda et al. \cite{Pano-Fukuda}} & \vcell{DetectNet}   & \vcell{240}& \vcell{0.748}  & \vcell{2019}   \\ [-\rowheight]
\printcelltop & \printcelltop   & \printcelltop  & \printcelltop  & \printcelltop  \\
 
\vcell{Lee et al. \cite{Pano-Lee-3}}& \vcell{CNN, VGG-16} & \vcell{544}& \vcell{0.840 ± 0.018 (VGG-16)} & \vcell{2020}   \\ [-\rowheight]
\printcelltop & \printcelltop   & \printcelltop  & \printcelltop  & \printcelltop  \\
 
\vcell{Bouchahma et al. \cite{Pano-Bouchahma}}  & \vcell{CNN} & \vcell{200 (ROI of teeth)} & \vcell{0.86}   & \vcell{2019}   \\ [-\rowheight]
\printcelltop & \printcelltop   & \printcelltop  & \printcelltop  & \printcelltop  \\
 
\vcell{Murata et al. \cite{Pano-Murata}} & \vcell{AlexNet} & \vcell{800 (ROIs)} & \vcell{0.875}  & \vcell{2019}   \\ [-\rowheight]
\printcelltop & \printcelltop   & \printcelltop  & \printcelltop  & \printcelltop  \\
 
\vcell{Chu et al. \cite{Pano-Chu}}& \vcell{AlexNet} & \vcell{108}& \vcell{0.8981} & \vcell{2018}   \\ [-\rowheight]
\printcelltop & \printcelltop   & \printcelltop  & \printcelltop  & \printcelltop  \\
 
\vcell{Oktay \cite{Pano-Oktay-2}} & \vcell{Stacked Auto-Encoders CNN}   & \vcell{10} & \vcell{0.91}   & \vcell{2018}   \\ [-\rowheight]
\printcelltop & \printcelltop   & \printcelltop  & \printcelltop  & \printcelltop  \\
 
\vcell{Sukegawa et al. \cite{Pano-Sukegawa}}   & \vcell{CNN, VGG-16, VGG-19} & \vcell{6645 (implant ROI)} & \vcell{0.935}  & \vcell{2020}   \\ [-\rowheight]
\printcelltop & \printcelltop   & \printcelltop  & \printcelltop  & \printcelltop  \\
 
\vcell{Singh et al. \cite{Pano-Singh}}  & \vcell{CNN} & \vcell{240}& \vcell{0.95}   & \vcell{2020}   \\ [-\rowheight]
\printcelltop & \printcelltop   & \printcelltop  & \printcelltop  & \printcelltop  \\\vcell{Jader et al. \cite{Pano-Jader}}  & \vcell{Mask R-CNN}  & \vcell{193}& \vcell{0.98 ± 0.008}   & \vcell{2018}   \\ [-\rowheight]
\printcelltop & \printcelltop   & \printcelltop  & \printcelltop  & \printcelltop  \\\vcell{Lee et al. \cite{Pano-Lee-1}}& \vcell{AlexNet} & \vcell{1068}   & \vcell{0.985}  & \vcell{2019}   \\ [-\rowheight]
\printcelltop & \printcelltop   & \printcelltop  & \printcelltop  & \printcelltop  \\
 
 \vcell{Lakshmi et al. \cite{Pano-Lakshmi}}& \vcell{AlexNet} & \vcell{900}& \vcell{0.986}  & \vcell{2020}   \\ [-\rowheight]
\printcelltop & \printcelltop   & \printcelltop  & \printcelltop  & \printcelltop  \\\vcell{Wirtz et al. \cite{Pano-Wirtz}}  & \vcell{U-Net}   & \vcell{10} & \vcell{0.744 (DSC)}& \vcell{2018}   \\ [-\rowheight]
\printcelltop & \printcelltop   & \printcelltop  & \printcelltop  & \printcelltop  \\
 \vcell{Chung et al. \cite{Pano-Chung}}  & \vcell{\begin{tabular}[b]{@{}l@{}}ResNet, \\  Deep Layer Aggregation,\\  Stacked-Hourglass \end{tabular}} & \vcell{574}& \vcell{0.84 (mIoU)}& \vcell{2020}   \\ [-\rowheight]
\printcelltop & \printcelltop   & \printcelltop  & \printcelltop  & \printcelltop  \\\vcell{Lee et al. \cite{Pano-Lee-4}}& \vcell{mask R-CNN}  & \vcell{30} & \vcell{0.879 (IoU)}& \vcell{2020}   \\ [-\rowheight]
\printcelltop & \printcelltop   & \printcelltop  & \printcelltop  & \printcelltop  \\
 
\vcell{Koch et al. \cite{Pano-Koch}}   & \vcell{U-Net}   & \vcell{1201}   & \vcell{0.936 (Dice)}   & \vcell{2019}   \\ [-\rowheight]
\printcelltop & \printcelltop   & \printcelltop  & \printcelltop  & \printcelltop  \\
 
\vcell{Vinayahalingam et al. \cite{Pano-Vinayahalingam}} & \vcell{U-Net}   & \vcell{57} & \vcell{\begin{tabular}[b]{@{}l@{}}0.936 (DSC - third molar)\\  0.805 (DSC - nerve) \end{tabular}}& \vcell{2019}   \\ [-\rowheight]
\printcelltop & \printcelltop   & \printcelltop  & \printcelltop  & \printcelltop  \\
 
\vcell{Oktay \cite{Pano-Oktay-1}} & \vcell{AlexNet} & \vcell{100}& \vcell{\begin{tabular}[b]{@{}l@{}}0.9432 (Dice - molar class)\\  0.9247 (Dice - premolar class)\\  0.9174 (Dice - canine incisor class) \end{tabular}}   & \vcell{2017}   \\ [-\rowheight]
\printcelltop & \printcelltop   & \printcelltop  & \printcelltop  & \printcelltop  \\
 
\vcell{Dasanayaka et al. \cite{Pano-Dasanayaka}} & \vcell{U-Net}   & \vcell{800}& \vcell{0.987 (DSC)}& \vcell{2019}   \\ [-\rowheight]
\printcelltop & \printcelltop   & \printcelltop  & \printcelltop  & \printcelltop  \\
 
\vcell{Du et al. \cite{Pano-Du}} & \vcell{CNN} & \vcell{5166 ROI}   & \vcell{MSE = 0.339}& \vcell{2018}   \\ [-\rowheight]
\printcelltop & \printcelltop   & \printcelltop  & \printcelltop  & \printcelltop  \\
 
\vcell{Ren et al. \cite{Pano-Ren}}& \vcell{RetinaNet, statistical shape model}  & \vcell{86} & \vcell{0.03816 (mean loss in pixels)}  & \vcell{2020}   \\ [-\rowheight]
\printcelltop & \printcelltop   & \printcelltop  & \printcelltop  & \printcelltop  \\
 
\vcell{Kim et al. \cite{Pano-Kim-2} }& \vcell{Faster R-CNN, Inception-V3}  & \vcell{253}& \vcell{0.845  (Precision)} & \vcell{2020}   \\ [-\rowheight]
\printcelltop & \printcelltop   & \printcelltop  & \printcelltop  & \printcelltop  \\
 
\vcell{Silva et al. \cite{Pano-Silva} }  & \vcell{Mask R-CNN, PANet, HTC, ResNeSt} & \vcell{324}& \vcell{71.3 ± 0.3 (mAP)}   & \vcell{2020}   \\ [-\rowheight]
\printcelltop & \printcelltop   & \printcelltop  & \printcelltop  & \printcelltop  \\
 
\vcell{Kwon et al. \cite{Pano-Kwon}}   & \vcell{YOLOv3}  & \vcell{946}& \vcell{\begin{tabular}[b]{@{}l@{}}0.88 ± 0.04 (Detection - mean AP)\\  0.956 ± 0.015 (Classification) \end{tabular}}& \vcell{2020}   \\ [-\rowheight]
\printcelltop & \printcelltop   & \printcelltop  & \printcelltop  & \printcelltop  \\
 \vcell{Kim et al. \cite{Pano-Kim-1}}& \vcell{U-Net}   & \vcell{11189}  & \vcell{0.87 (sensitivity)} & \vcell{2019}   \\ [-\rowheight]
\printcelltop & \printcelltop   & \printcelltop  & \printcelltop  & \printcelltop  \\
 \vcell{Muramatsu et al. \cite{Pano-Muramatsu}}  & \vcell{DetectNet, ResNet-50}& \vcell{75} & \vcell{\begin{tabular}[b]{@{}l@{}}0.964 (detection sensitivity)\\  0.932 (tooth type classification)\\  0.98 (tooth condition classification) \end{tabular}} & \vcell{2020}   \\ [-\rowheight]
\printcelltop & \printcelltop   & \printcelltop  & \printcelltop  & \printcelltop  \\
\vcell{Tuzoff et al. \cite{Pano-Tuzoff}} & \vcell{Faster R-CNN, VGG-16}& \vcell{1352}   & \vcell{\begin{tabular}[b]{@{}l@{}}0.9941 (sensitivity of detection)\\  0.9800 (sensitivity of numbering) \end{tabular}}  & \vcell{2019}   \\ [-\rowheight]
\printcelltop & \printcelltop   & \printcelltop  & \printcelltop  & \printcelltop  \\
\bottomrule
\end{tabular*}
\end{table*}

Works related to panoramic x-ray images of teeth fall mainly into 4 categories: 1) classification, 2) detection, 3) treatment prediction, and 4) segmentation.

\newpage
\textbf{Classification} Oktay \cite{Pano-Oktay-1} applies a AlexNet-inspired architecture to classify between 3 types of teeth and the background. Next, the same author published a paper \cite{Pano-Oktay-2} which concerns placing a landmark on each tooth. For this, they developed a stacked auto-encoder to narrow down on a ROI to maintain a high resolution. Murata et al. \cite{Pano-Murata} use an AlexNet architecture to classify a ROI of a maxillary sinus as normal or inflamed from panoramic x-rays. The deep learning system achieves better results than the dental residents, but worse than radiologists, on the test set. These works use AlexNet or encoders for classification while the following works use their own or VGG-based architectures.
Singh et al. \cite{Pano-Singh} create their own 6-layer CNN structure to number and classify teeth. They use 240 panoramic x-rays and automatically extract ROIs of teeth. It is unclear if a human validated that each tooth has a proper bounding box and number. Sukegawa et al. \cite{Pano-Sukegawa} compares their own CNN, VGG-16, and VGG-19 for the classification of models of implants. With 11 classes and a dataset of 6645 implant patches, they obtained an average accuracy of 0.935.

\textbf{Detection} 
Tuzoff et al. \cite{Pano-Tuzoff} method is a two-stage process for detecting teeth and numbering them on a scale of 1 to 32. First, a Faster R-CNN network detects them and then a VGG-16 classifies them into a number. The paper presents interesting edge-cases where the model is correct and the expert's labels incorrect. Their method is evaluated using the sensitivity metric. Sensitivity aka recall is a ratio of how many positives samples were picked from all available positive samples (see Eq. \ref{eq:accuracy}). It is represented with the following equation: 
\begin{equation}
    \frac{TP}{TP+FN},
    \label{eq:sensitivity}
\end{equation}
where $TP$ and $FN$ are True Positive, False Negative respectively.
Chung et al. \cite{Pano-Chung} study the task of detecting and numbering teeth on an x-ray. Their paper compares different methods (Faster R-CNN, CenterNet \cite{CenterNet} or their proposed method) with different backbones (ResNet, Deep Layer Aggregation \cite{DeepLayerAggregation} or Stacked-Hourglass \cite{StackedHourglass}). 
Kim et al. \cite{Pano-Kim-2} apply Faster R-CNN to the task of tooth detection and numbering. The model omits dislocated teeth and residual teeth. They expand the dataset to include implants, while earlier works omitted them from the dataset. This method uses precision as one of its evaluation metrics. Precision is a measure of how many true positives are in the positive results return by the model (see Eq. \ref{eq:precision}). It is calculated through the following equation: 
\begin{equation}
    \frac{TP}{TP+FP},
    \label{eq:precision}
\end{equation}
where $TP$ and $FP$ are True Positive and False Positive respectively. Chung et al. and Kim et al. both use a Faster RCNN for comparison. Kim et al. \cite{Pano-Kim-2} achieved 0.967 mAP at IoU = 0.5, which is higher than both the Faster R-CNN or method proposed by Chung et al. \cite{Pano-Chung}. The IoU metric, aka Jacquard index, is measure of similarity between the predicted results and ground truth (see Eq. \ref{eq:jaccard}). It is measured with the following equation:
\begin{equation}
    J(X, Y) = \frac{\lvert X \cap Y \rvert}{\lvert X \cup Y \rvert},
    \label{eq:jaccard}
\end{equation}
where $X$ and $Y$ are sets of machine predicted and ground truth pixels respectively. Mean Average Precision (mAP) is the average value over selected points on a recall vs. precision graph with an IoU value at or above 0.5.

The cause for this may be a difference in backbone architectures (Inception-V3 vs. ResNet18). Fukuda et al. \cite{Pano-Fukuda} use DetectNet \cite{DetectNet} to detect teeth with Vertical Root Fracture. Ren et al. \cite{Pano-Ren} attempt to find 8 trabecular landmarks. Their statistical shape model obtains a smaller loss than RetinaNet \cite{RetinaNet} (0.03816 vs. 0.0458 MRE). Kwon et al. \cite{Pano-Kwon} train a YOLOv3 architecture to detect and classify cysts and tumors in an x-ray. Their dataset is relatively large but biased, as normal jaws accounted for around ~10\% of the dataset. Muramatsu et al. \cite{Pano-Muramatsu} used DetectNet. They show a multisize network that has 2 inputs: an image with a proper bounding box and an image with 2x the width to include the neighboring teeth. The wider ROI made it easier for the network to properly classify a tooth.

\textbf{Treatment prediction} 
Chu et al. \cite{Pano-Chu} build a siamese network that incorporates AlexNet for the classification of panoramic x-rays as having osteoporosis or being normal. Lee et al. \cite{Pano-Lee-1} use AlexNet for the classification of osteoporosis. Splitting an ROI into two smaller boxes and concatenating the output in the network gave better results than having a single ROI as the input. However, their dataset ground-truth label is set by dental specialists, not by the T-Score, the gold standard in the industry. Lee et al. \cite{Pano-Lee-3} also research the problem of classifying osteoporosis. Their experiments show that for the VGG-16 and VGG-19 architectures, transfer learning can be applied to medical images to improve results.
Lee et al. \cite{Pano-Lee-1} method with AlexNet achieved the best result of the three, with a classification accuracy of 0.985. Although they used a similar structure as Chu et al. \cite{Pano-Chu} work, they had 10x the training data. The difference in results between  Lee et al. \cite{Pano-Lee-1} and Lee et al. \cite{Pano-Lee-3} of 0.985 vs. 0.84 $\pm$ 0.018 may be due to a higher ground truth standard for the latter work.
Du et al. \cite{Pano-Du} use a CNN to reposition dental arches on x-rays and reconstruct them to minimize blur. They obtain results of 0.339 MSE distance between their output and ground truth. However, the results could be biased as only 1\% of the dataset was used as a test set and as a validation set each, with the remaining data (98\%) being used for training.
Lakshmi et al. \cite{Pano-Lakshmi} classify teeth as 'decay' or 'normal'. They use an AlexNet-like architecture with a conventional approach of training the model on their ROI’s. Bouchahma et al. \cite{Pano-Bouchahma} create a CNN to classify if a tooth did not need treatment, needed a fluoride treatment, a filling, or a root canal.

\textbf{Segmentation} Wirtz et al. \cite{Pano-Wirtz} use U-Net to initialize a statistically shaped model for tooth segmentation. Koch et al. \cite{Pano-Koch} use a U-Net architecture for semantic segmentation on the open dataset from \cite{Pano-Silva}. They are able to improve the Dice score from 0.744 obtained with Wirtz et al. \cite{Pano-Wirtz} method to 0.936. Dice's coefficient aka Dice similarity coefficient (DSC) is a measure of similarity between the predicted results and ground truth (see Eq. \ref{eq:dsc}). The equation is the following: 
\begin{equation}
    DSC = \frac{2 \lvert X \cap Y  \rvert}{\lvert X \rvert + \lvert Y \rvert},
    \label{eq:dsc}
\end{equation}
where $X$ and $Y$ would be sets of machine segmented and ground truth pixels respectively. \cite{Yeghiazaryan}. One factor which helps the method achieve a higher Dice score is that they have annotated more images from the dataset than the previous study. Vinayahalingam et al. \cite{Pano-Vinayahalingam} use a U-Net architecture to segment third molars and the mandibular nerve. Third molars (aka. Wisdom teeth) are frequently excluded from datasets due to limited data points. They obtain a similar DSC score of 0.936 as with the standard teeth from the earlier study.
Dasanayaka et al. \cite{Pano-Dasanayaka} used a U-Net architecture to segment two openings in the mandible called the mental foramen with a result of 0.987 DSC.
Kim et al. \cite{Pano-Kim-1} use a two-stage U-net architecture to segment periodontal bone loss. Their model has a better AUC, F1 score, Specificity, and PPV as compared to dentistry residents. However, the residents perform better on Sensitivity, NPV, and third molars.
Jader et al. \cite{Pano-Jader} apply an architecture based on Mask R-CNN \cite {maskrcnn} for segmentation. Mask R-CNN is a two-stage network built on top of ResNet. The first stage generates proposals. The second stage selects the proposals and generates masks. They show a breakdown of their dataset, with variations such as all/missing teeth, with/without restoration, with/without a dental appliance, with implants, or with more than 32 teeth. They use the open dataset from \cite{Pano-Silva} but annotate and use only a subset of 193 images from the available 1500 images. With more images of different variants, the model is more generalizable than Wirtz et al. \cite{Pano-Wirtz}.
Lee et al. \cite{Pano-Lee-4} apply the Mask R-CNN architecture to the problem of teeth segmentation. Comparing their method to \cite{Pano-Wirtz}, they have a better Precision score but it is unknown if they train the model on their dataset or copy the results from the paper, as Lee et al.’s paper does not use the open dataset.

The diagram Fig. \ref{fig:panoramic_xray} shows joint segmentation and classification in an X-ray panoramic images using convolutional neural networks.

Silva et al. \cite{Pano-Silva} work compared SOTA segmentation methods on an additionally annotated version of their open dataset. They compared the results for Mask R-CNN, PANet, HTC, and ResNeSt. PANet had the best results, which was surprising as both of the previous papers use Mask R-CNN for segmentation.

The Table \ref{tab:panoramic_table} shows a comparison of works that use X-ray panoramic images.

\subsection{Cone beam computed tomography (CBCT)}

\begin{figure*}
    \centering
    \includegraphics[width=14.0cm]{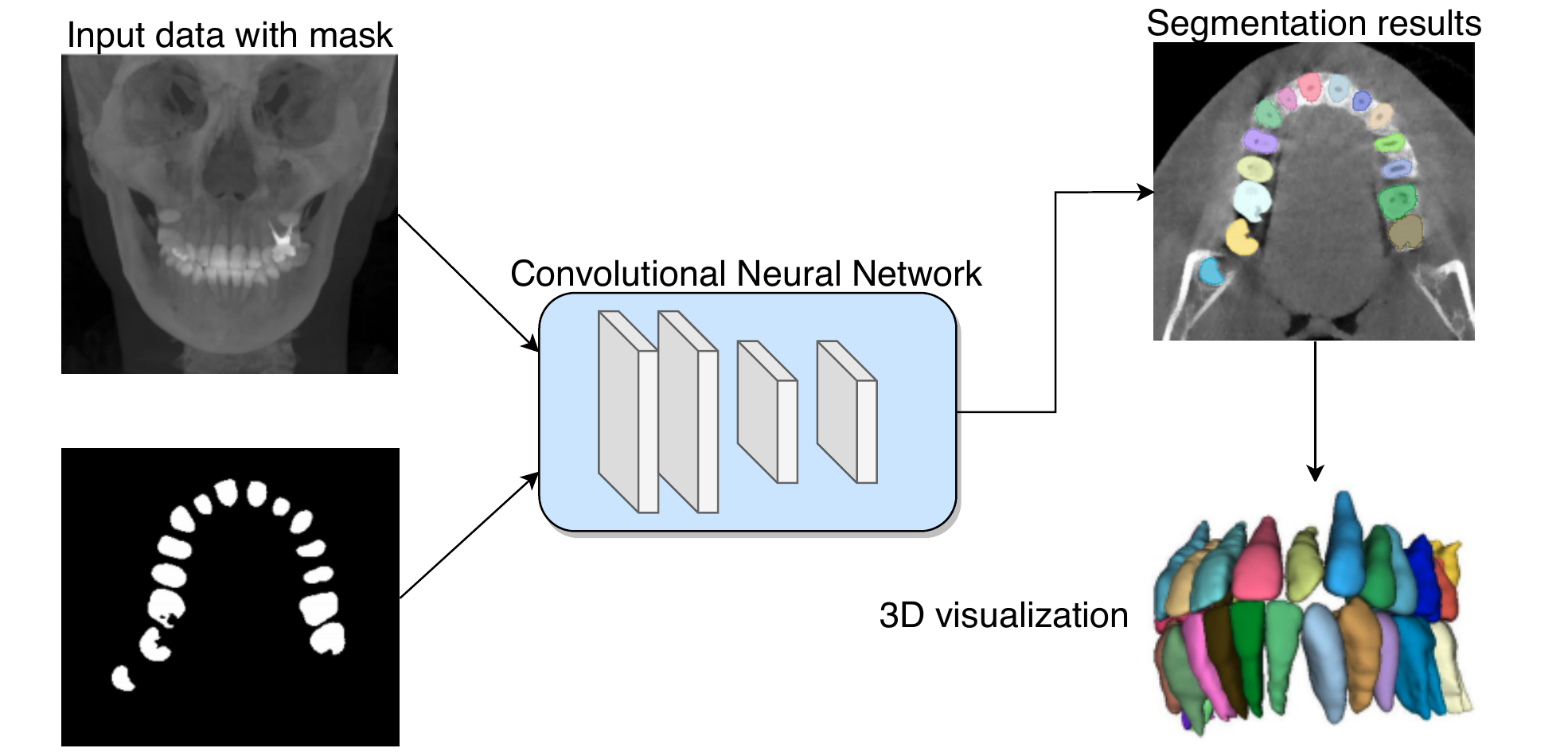}
    \caption{An example of a tooth segmentation scheme from 3D CBCT images. From left: 3D CBCT image and masks, convolutional neural network (e.g. U-Net), output segmentation and 3D visualization}
    \label{fig:cbct}
\end{figure*}

\begin{figure}[ht!]
\begin{minipage}[b]{.49\linewidth}
  \centering
  \centerline{\includegraphics[width=4.0cm, height=3.5cm]{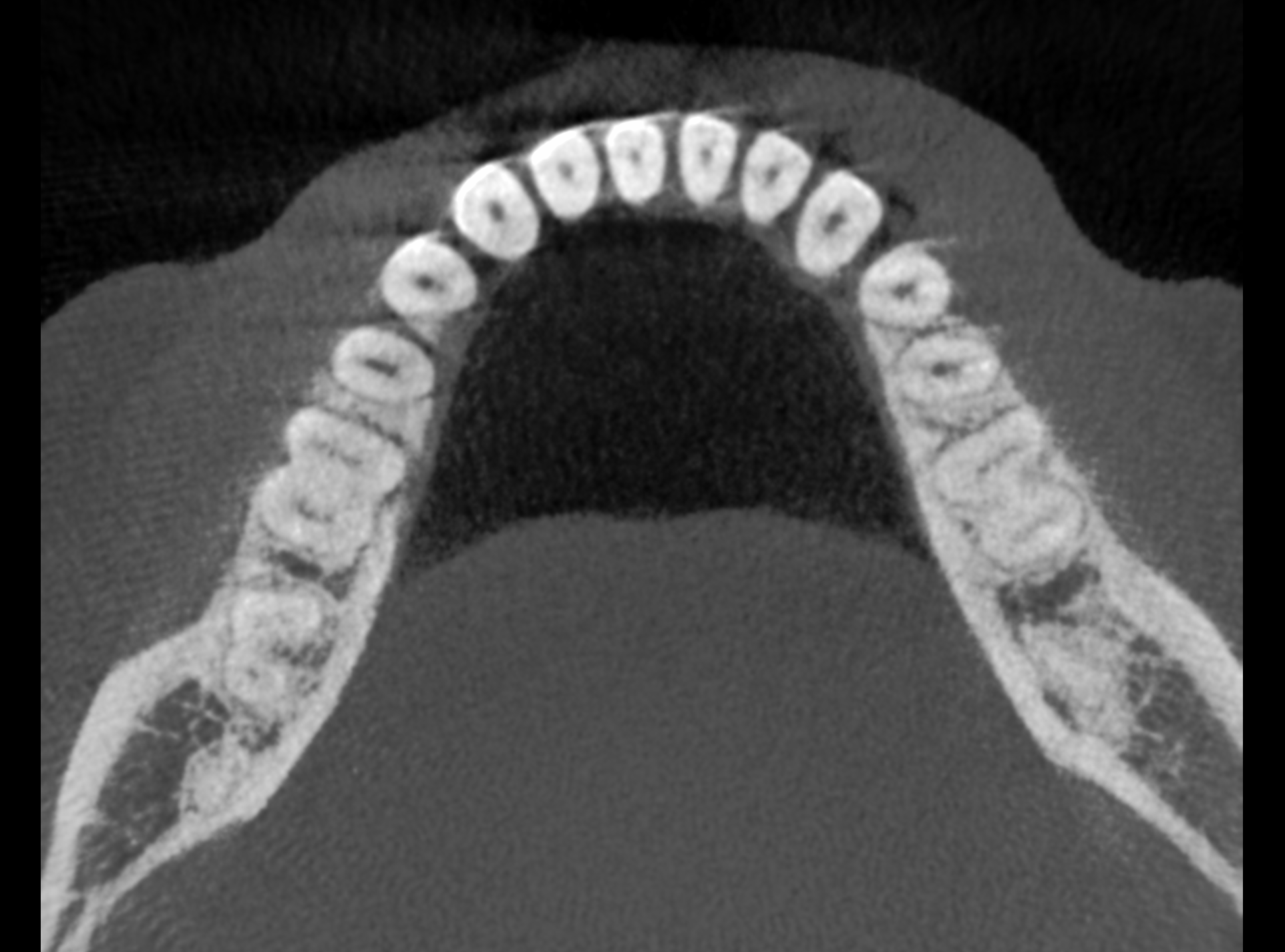}}
\end{minipage}
\hfill
\begin{minipage}[b]{0.49\linewidth}
  \centering
  \centerline{\includegraphics[width=4.0cm, height=3.5cm]{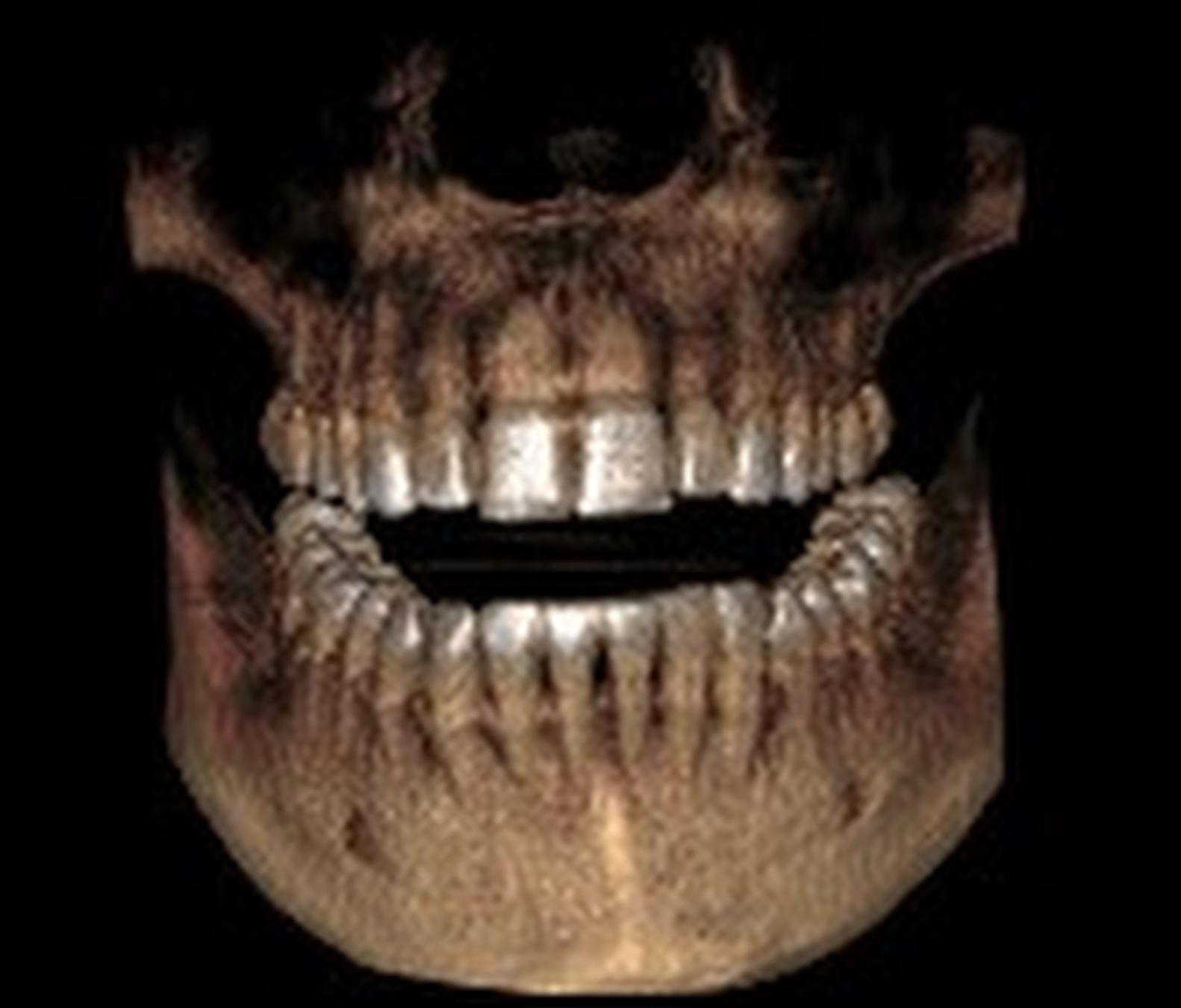}}
\end{minipage}
\caption{Example of Cone-Beam Computed Tomography (CBCT)}
\label{fig:cbct_dataset}
\end{figure}

Cone Beam Computed Tomography (CBCT) images are created by using a flat panel sensor and projection beam to capture 2D images, which are then mathematically converted to a 3D image. They allow for the creation of a 3D model of the patient's skull (see Fig. \ref{fig:cbct_dataset}). Due to its high resolution, CBCT imaging is extremely useful in orthodontics. CBCT is used in planning the location of orthodontic implants and micro-implants, in determining the location of impacted, supernumerary and additional teeth, in the assessment of root resorption, and the assessment of the thickness of the bone tissue of the maxilla and the alveolar part of the mandible.
 
Moreover, this examination allows for the imaging of the temporomandibular joint and is useful in the treatment planning of patients with congenital defects of the facial part of the skull. CBCT scans were introduced to dental practices in the United States between 2001 and 2004 \cite{Larson}. First digital tooth segmentation methods date from 2005 \cite{Zhao} and include techniques such as morphological operators \cite{Kakehbaraei}, marker-based watershed algorithms \cite{Kakehbaraei,Fan}, region growing \cite{Akhoondali}, template-based methods \cite{Barone, Pei}, graph-cut-based approaches \cite{Hiew} and random forests \cite{Wang}.

These classical methods are brittle and require fine-tuned heuristics or shape priors. Data-intensive methods like CNNs can handle a wider range of CBCT scans with better results. Works related to Cone-Beam Computed Tomography (similarly to works related to panoramic X-ray) can be divided into two groups according to their application: classification and segmentation.

\begin{table*}[width=\textwidth, cols=4, pos=h]
\centering
\caption{Comparison of papers based on Cone-Beam Computed Tomography (CBCT) images}
\label{tab:cbct_table}
\begin{tabular*}{\tblwidth}{@{} LLLL@{} }
\toprule
Author &  Architecture&  Evaluation (average accuracy if not mentioned)   &  Year   \\
\midrule
\vcell{Pavaloiu et al. \cite{Pavaloiu}}   & \vcell{Neural Network}  & \vcell{Quantitative}  & \vcell{2015}   \\[-\rowheight]
\printcelltop & \printcelltop   & \printcelltop & \printcelltop  \\
\vcell{Miki et al. \cite{MikiB}}   & \vcell{CNN (AlexNet)}   & \vcell{0.744 (detection)} & \vcell{2017}   \\[-\rowheight]
\printcelltop & \printcelltop   & \printcelltop & \printcelltop  \\
\vcell{Miki et al. \cite{MikiA}}   & \vcell{CNN (AlexNet)}   & \vcell{0.888} & \vcell{2017}   \\[-\rowheight]
\printcelltop & \printcelltop   & \printcelltop & \printcelltop  \\
\vcell{Egger et al. \cite{Egger}}  & \vcell{CNN (VGG-16, FCN)}   & \vcell{0.8964}& \vcell{2018}   \\[-\rowheight]
\printcelltop & \printcelltop   & \printcelltop & \printcelltop  \\
\vcell{Lee et al. \cite{Lee-2p}}& \vcell{Inception-V3}& \vcell{\begin{tabular}[b]{@{}l@{}}0.914 (CBCT accuracy),\\ 0.846 (Panoramic X-ray accuracy) \end{tabular}}& \vcell{2020}   \\[-\rowheight]
\printcelltop & \printcelltop   & \printcelltop & \printcelltop  \\
\vcell{Kim et al. \cite{Kim}}& \vcell{VGG-16 or Inception-V3}  & \vcell{\begin{tabular}[b]{@{}l@{}}0.9333 (VGG-16),\\ 0.9383 (Inception-V3) \end{tabular}} & \vcell{2020}   \\[-\rowheight]
\printcelltop & \printcelltop   & \printcelltop & \printcelltop  \\
\vcell{Kwak et al. \cite{Kwak}}   & \vcell{2D U-Net, 3D U-Net and 2D SegNet}& \vcell{\begin{tabular}[b]{@{}l@{}}0.57721 (3D U-Net, mean IoU),\\ 0.49116 (2D SegNet, mean IoU),\\ 0.45984 (2D U-Net, mean IoU), \\ 0.95915 (3D U-Net, class average accuracy),\\ 0.90271 (2D SegNet, class average accuracy),\\ 0.68310 (2D U-Net, class average accuracy) \end{tabular}} & \vcell{2020}   \\[-\rowheight]
\printcelltop & \printcelltop   & \printcelltop & \printcelltop  \\
\vcell{Setzer et al. \cite{Setzer}} & \vcell{U-Net}   & \vcell{0.714 (Dice, average from all labels)} & \vcell{2020}   \\[-\rowheight]
\printcelltop & \printcelltop   & \printcelltop & \printcelltop  \\
\vcell{Chung et al. \cite{Chung}}  & \vcell{2D CNN + Faser R-CNN + 3D U-Net} & \vcell{\begin{tabular}[b]{@{}l@{}}0.86 ± 0.01 mm (Jaccard index, tooth detection),\\ 0.20 ± 0.10 mm (ASSD, tooth detection), \\ 0.15 ± 0.0.04 mm (ASSD, instance segmentation) \end{tabular}} & \vcell{2020}   \\[-\rowheight]
\printcelltop & \printcelltop   & \printcelltop & \printcelltop  \\
\vcell{Minnema et al. \cite{Minnema}}& \vcell{MS-D CNN}& \vcell{0.87 ± 0.06 (DSC)} & \vcell{2019}   \\[-\rowheight]
\printcelltop & \printcelltop   & \printcelltop & \printcelltop  \\
\vcell{Lee et al. \cite{SLee}}& \vcell{U-Net}   & \vcell{0.918 (Dice)}  & \vcell{2020}   \\[-\rowheight]
\printcelltop & \printcelltop   & \printcelltop & \printcelltop  \\
\vcell{Cui et al. \cite{Cui}}& \vcell{3D Mask R-CNN}   & \vcell{\begin{tabular}[b]{@{}l@{}}0.9237 (DSC), 0.9955 (detection accuracy), \\ 0.9582 (identification accuracy) \end{tabular}}   & \vcell{2019}   \\[-\rowheight]
\printcelltop & \printcelltop   & \printcelltop & \printcelltop  \\
\vcell{Chen et al. \cite{Chen}}   & \vcell{\begin{tabular}[b]{@{}l@{}}V-Net + \\Marker-controlled watershed transform \end{tabular}} & \vcell{\begin{tabular}[b]{@{}l@{}}0.936 ± 0.012 (DSC), \\ 0.881 ± 0.019 (Jaccard index), \\ 0.072 ± 0.027 (relative volume difference), \\ 0.363 ±0.145 mm (ASSD) \end{tabular}}  & \vcell{2020}   \\[-\rowheight]
\printcelltop & \printcelltop   & \printcelltop & \printcelltop  \\
\vcell{Torosdagli et al. \cite{Torosdagli} } & \vcell{CNN (Tiramisu, U-Net),  LSTM}& \vcell{0.9382 (DSC)}  & \vcell{2019}   \\[-\rowheight]
\printcelltop & \printcelltop   & \printcelltop & \printcelltop  \\
\vcell{Ezhov et al. \cite{Ezhov}}  & \vcell{V-Net}   & \vcell{0.94 (IoU), 0.17 mm (ASSD)} & \vcell{2018}   \\[-\rowheight]
\printcelltop & \printcelltop   & \printcelltop & \printcelltop  \\
\vcell{Wu et al. \cite{Wu}} & \vcell{U-Net}   & \vcell{\begin{tabular}[b]{@{}l@{}}0.962 (DSC),\\ 0.995 (detection accuracy),\\ 0.991 (identification accuracy),\\ 0.122 (ASSD) \end{tabular}}  & \vcell{2020}   \\[-\rowheight]
\printcelltop & \printcelltop   & \printcelltop & \printcelltop \\
\bottomrule
\end{tabular*}
\end{table*}

\textbf{Classification} Pavaloiu et al. \cite{Pavaloiu} design a 2 layer network that performs edge detection of teeth on a CBCT axial slice. There is no prominent architecture and no quantitative results for comparison with future methods. Miki et al. \cite{MikiA} use AlexNet to classify a ROI from a CBCT axial slice into 7 tooth types. They omit ROIs of the 3rd molar or metal artifacts.
In their next paper, Miki et al. \cite{MikiB} modify AlexNet to output a heatmap of a CBCT axial slice for tooth detection and apply bounding boxes on the result if there is over a 95\% confidence that it is a tooth. Kim et al. \cite{Kim} present a method to classify CBCT scans into three types of malocclusions based on the analysis of 2D projections. Their experiments compare pre-trained VGG-16- and Inception-V3-based architectures for classification. VGG-16 based architecture has the best results when the results from 3 projected images were chosen based on a voting scheme. The Inception-V3 model receives the best results when the output was based on a two-step learning method described in the paper. Lee et al. \cite{Lee-2p} present work on the classification of cystic lesions from CBCT axial scans and panoramic x-rays. They use Inception-V3 with transfer learning for classification. CBCT scans were easier to classify accurately than panoramic images.

\textbf{Segmentation-2D} Egger et al. \cite{Egger} use CNNs for mandible segmentation. First, they use a VGG-16 classification network to check if the image contains the mandible or not. If it does, it passes it through three VGG-16-based nets for the segmentation of the mandible. They achieve a Dice score of 0.8964 and confirm that a larger dataset would lead to a better result. Minnema et al. \cite{Minnema} present a mixed-scale dense CNN for metal artifact reduction. Their method has better results than the clinical benchmark, snake evolution. They receive comparable results to U-Net and ResNet with 300x and 700x less trainable parameters respectively.  Subsequent works are based on one of the most popular neural network architectures used for segmentation - U-Net. Torosdagli et al. \cite{Torosdagli} propose a method for automatic segmentation and annotation of 9 anatomical landmarks on the mandible. Their three-step method is composed of a Tiramisu network for segmentation, a U-Net-based learning algorithm, and finally an LSTM network. In addition to their private dataset, they use the publicly available MICCAI Head-Neck Challenge (2015) \cite{miccai-head-neck-challenge} dataset for comparison. Kwak et al. \cite{Kwak} research concerns the task of segmenting the mandibular canal in a CBCT scan. They compare different methods based on 2D U-Net, 3D U-Net \cite{3dunet}, and 2D SegNet \cite{segnet}. Interestingly, SegNet shows better results than the widely-used U-Net. The 3D U-Net has the best results, however, the image is downsized to 132 x 132 x 132, which results in a loss of accuracy. Lee et al. \cite{SLee} also propose a method for tooth segmentation from a CBCT scan based on U-Net architecture. To allow the model to converge faster but also be able to handle empty voxels, they set up a multi-phase training method, with each phase increasing the area around the teeth which is passed as the training dataset. Their method can handle metal artifacts and loss of teeth well. Setzer et al. \cite{Setzer} apply U-Net architecture for the detection of periapical lesions in CBCT scans. The network output has 5 classes: lesion, tooth, bone, restorative materials, and background. Minnema et al. \cite{Minnema} and Lee et al. \cite{SLee} both segment teeth from a CBCT scan. Lee et al. \cite{SLee} obtains better results (0.917 vs. 0.87 DSC) with more training data and had more, precisely 73, scans with metal artifacts. However, the MS-D network \cite{Minnema} can achieve those results with much less training parameters.

\textbf{Segmentation-3D} Chen et al. \cite{Chen} and Ezhov et al. \cite{Ezhov} propose methods which use a V-Net \cite{mia1}.
The Chen et al. \cite{Chen} method uses a V-Net to output a tooth probability map and a tooth surface map. This is inputted into a marker-controlled watershed transform for tooth segmentation. They show that a patch size of 64 is optimal for their segmentation method. They have a small dataset due to the time requirement of making voxel-level masks. Ezhov et al. method \cite{Ezhov} presents results for the segmentation of CBCT voxels into 33 classes (32 teeth + background). The method consists of two sequential models: the coarse model which outputs 33 classes and the fine model which outputs 2 classes: if the data is of a given tooth type or not. They also use 2 types of datasets: coarse and precise. The coarse dataset is created with linearly interpreted bounding boxes of CBCT axial slices. The fine dataset is a per voxel mask created manually. The paper shows that using a coarse dataset has an improved result compared to a fine dataset only. Due to the time requirement of having a precise dataset, there were only 120 precise scans. The method used Average Symmetric Surface Distance as one of the metrics to evaluate the model. Average Symmetric Surface Distance aka ASSD or ASD is the average distance from all points on the boundary of the machine predicted mask to ground truth mask and vice versa \cite{Yeghiazaryan}. Cui et al. \cite{Cui} present the ToothNet architecture in their paper, based on Mask R-CNN. The two-step method consists of an edge detection step of the CBCT followed by a 3D region proposal module. Their method performs well on 3D data both quantitatively and qualitatively. Chung et al. \cite{Chung} method for segmentation improved on ToothNet to include metal artifacts. They use a 2D CNN for pose regression, then a Faster R-CNN based network for tooth detection, and finally a 3D U-Net based network for tooth segmentation. 

Compared to ToothNet, they obtain a 0.93 F1 score while ToothNet scores 0.88. F1-score is a harmonic mean of recall and precision combined (see Eq. \ref{eq:f1}) in the following way: 
\begin{equation}
    2 * \frac{precision*recall}{precision+recall}
    \label{eq:f1}
\end{equation}
It is used to compress the performance of a model into a single metric. Wu et al. \cite{Wu} also tackle the problem of tooth instance segmentation. They propose a two-stage framework. The first stage is a heatmap regression U-Net for tooth ROIs. The second stage is a network based on U-Net and Spatial Pyramid Pooling for segmentation. They have a small dataset due to the time requirement of making voxel-level masks. The last work, by using heatmap regression and Pyramid Polling \cite{DenseASPP} can retain the resolution of ROI and obtain better results with a DSC of 0.962 than either of the previous papers.
Example of a diagram (see Fig. \ref{fig:cbct}) that shows Cone-Beam Computed Tomography (CBCT) processing to segment teeth.

Overall, initial papers were interested in tooth detection, then moved on to tooth segmentation. In 2018, the applications of CNN’s grew with initial papers in the fields of pathology detection based on CBCT data being published. ToothNet from 2019 was the first published paper concerning instance segmentation of teeth in CBCT images and was referenced by future papers which improved the proposed model to include wisdom teeth and dealing with metal artifacts.

We can see a large increase in papers published in 2020. These included state-of-the-art results in tooth instance segmentation (0.962 Dice), as well as the introduction of new research directions such as classification of craniofacial images, mandibular canal segmentation, detection of cysts, and lesions.

The Table \ref{tab:cbct_table} shows a comparison of works that use Cone-Beam Computed Tomography (CBCT) images.

\subsection{Dental casts}

\begin{figure*}[ht!, width=\textwidth]
    \centering
    \includegraphics[width=14.0cm]{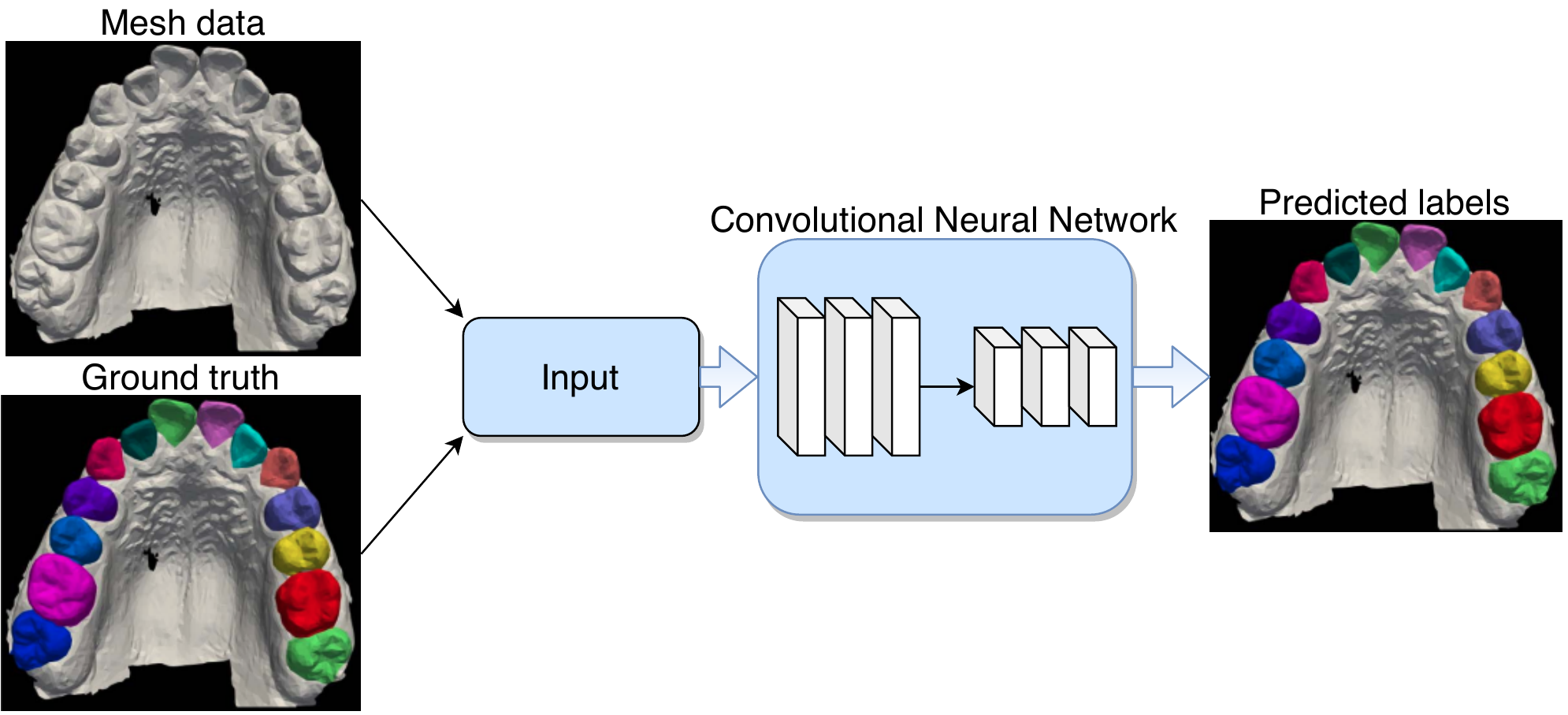}
    \caption{An example of a tooth segmentation scheme for 3D dental casts. From the left: 3D models in the form of mesh points and tooth annotations transferred to the convolutional network, with predicted labels as a result}
    \label{fig:dental_casts}
\end{figure*}

\begin{table*}[width=\textwidth, cols=4, pos=h]
\centering
\caption{Comparison of papers based on 3D dental casts models}
\label{tab:dentalcast}
\begin{tabular*}{\tblwidth}{@{} LLLL@{} }
\toprule
Author & Architecture   & Evaluation (average accuracy if not mentioned) & Year   \\
\midrule
\vcell{Raith et al. \cite{Raith}}   & \vcell{Neural Network} & \vcell{0.933}  & \vcell{2017}   \\[-\rowheight]
\printcelltop  & \printcelltop  & \printcelltop  & \printcelltop  \\
\vcell{Xu et al. \cite{Xu} }  & \vcell{CNN}& \vcell{0.9906} & \vcell{2018}   \\[-\rowheight]
\printcelltop  & \printcelltop  & \printcelltop  & \printcelltop  \\
\vcell{Lian et al. \cite{Lian}}& \vcell{CNN}& \vcell{\begin{tabular}[b]{@{}l@{}}0.938 (DSC)\\ 0.946 (SEN)\\ 0.934 (positive prediction value) \end{tabular}} & \vcell{2019}   \\[-\rowheight]
\printcelltop  & \printcelltop  & \printcelltop  & \printcelltop  \\
\vcell{Zanjani et al. \cite{Zanjani} } & \vcell{CNN}& \vcell{0.98 (IoU)} & \vcell{2019}   \\[-\rowheight]
\printcelltop  & \printcelltop  & \printcelltop  & \printcelltop  \\
\vcell{Kim et al. \cite{Kim2}} & \vcell{CNN (GAN)}  & \vcell{\begin{tabular}[b]{@{}l@{}}0.04 mm (improvement in comparison to\\ manual segmentation) \end{tabular}}  & \vcell{2020}   \\[-\rowheight]
\printcelltop  & \printcelltop  & \printcelltop  & \printcelltop  \\
\bottomrule
\end{tabular*}
\end{table*}

In addition to clinical examination and evaluation of radiographs, an important diagnostic criterion is the analysis of orthodontic models. This analysis is based on the assessment of the dental arch of the maxilla and mandible along three spatial planes. The mutual relations of the anterior and posterior width of the dental arches are assessed. Also, canine classes, Angle classes, as well as vertical and horizontal overlap, are specified. Gypsum models are not only a diagnostic tool but also constitute an important part of the orthodontic patient's documentation. They are frequently used even though they have several storage and damage problems \cite{Abizadeh}. One solution to these problems is using digital diagnostic models that were introduced in the 1990s \cite{Joffe, Mah}. Digital models have been shown to have the same value as plaster models and appear to have sufficient reliability for use in a clinical setting \cite{Dalstra}. However, the disadvantage of digital models is the difficulty in obtaining a reliable patient occlusion, which depends on the jaw relationship that the patient has acquired during the scan, while on plaster models it is recorded by an occlusion, collected and controlled by a doctor. Works related to dental casts of teeth fall mainly into one category: segmentation.

Digital model analysis is gaining increasing importance in orthodontics. CNNs have been employed to solve various problems related to teeth segmentation and analysis of digital models as well. Raith et al. \cite{Raith} present a method to classify dental cusps with sufficient accuracy using an Artificial Neural Network (ANN). Their method is validated on 129 models, consisting of 69 models of upper jaws and 60 models of lower jaws. They apply the so-called blob detection algorithm with a Difference of Gaussians (DoG) approach, as a feature detection step to the previously generated depth range images based on the acquired 3D surface data. Then, this prepared dataset was used for training and validation of the tailored ANN. They prove that an approach with ANNs shows high performance with correct classifications of 93.3\%. However, the proposed algorithm is not verified against unusual tooth geometry. Thus, additional research towards orthodontic patients should be considered for subsequent analysis, as the visualization of occlusal contacts is also important for orthodontic treatment.
Subsequent works are based on raw data from digital models and use them for tooth segmentation.
Xu et al. \cite{Xu} exploit deep convolutional neural networks (CNNs) for the task of 3D dental model segmentation. They extracted a set of geometry features by applying a boundary-aware mesh simplification method and a correspondence-free mapping algorithm to enable efficient feature extraction. Their model achieves a practical precision of 99.06\%.
A similar approach is used by Lian et al.\cite{Lian} and Zanjani et al.\cite{Zanjani}, which uses the architectural ideas of PointNet \cite{pointnet} and PointCNN \cite{pointcnn}, respectively.
Lian et al. \cite{Lian} propose a deep neural network (MeshSNet) to extract geometric features directly from the raw 3D dental surface for automated tooth segmentation. The result of segmentation is quantitatively evaluated by the following metrics: Dice similarity coefficient (DSC), sensitivity (SEN), and positive prediction value (PPV). MeshSNet achieves segmentation results of 0.938, 0.946, 0.934 for DSC, SEN, and PPV, respectively.
Zanjani et al. \cite{Zanjani} propose a solution called Mask-MCNet, as an analogy to Mask R-CNN \cite{maskrcnn}, which gives high performance in 2D images. The architecture is based on the Point CNN network, to which it adds a module based on Monte Carlo ConvNet (MCCNet) \cite{montecarlocnn}. In this work, the authors use the above method to segment teeth from 3D models, which was in the form of point clouds generated by intra-oral scanners. State-of-the-art results are obtained, reaching an IoU of 0.98.

An interesting approach in the segmentation process is the use of Generative Neural Networks (GANs) proposed by Kim et al. \cite{Kim2}. GANs consists of a generator and critic. The generator generates synthetic images and the critic has to discriminate if it is a real image or a synthetic one. This leads to a feedback loop where the synthetic images and the critic improve. Kim et al. attempt to solve the problem of occlusion and lack of accurate scan data in narrow interdental spaces. They prepare a reconstruction process of occluded areas by employing a Generative Adversarial Network (GAN). For the tooth segmentation task, they obtain results that prove an average improvement of 0.04 mm.
Example of a diagram (see Fig. \ref{fig:dental_casts}) that shows the processing of mesh data on 3D dental casts for tooth segmentation.
Table \ref{tab:dentalcast} shows a comparison of works that use 3D dental casts models.

\section{Future of CNNs in Orthodontics and Their Challenges}

The introduction in 1997 of clear aligner appliances had a great influence on future orthodontics development - these appliances take into account individual tooth anatomy and can perform tooth movement with optimal forces safe for the periodontium \cite{feature2}. Tooth movements are more physiological and the obtained treatment effect is more individually adjusted to the shape of the dental arches and the surrounding soft tissues. In the same way, using CNNs in orthodontic practice could lead to a similar improvement by taking into account individual tooth anatomy.
The development of technology has made orthodontics one of the fastest-growing branches of dentistry \cite{future1}. The development of three-dimensional imaging equipment and image processing methods is possible due to the large funding from the private sector. A correct bite has a positive effect on the functions of speech, chewing, swallowing, and breathing. In the case of an incorrect bite, the sense of self-confidence decreases, and patients are less successful in their professional and familial fields. Disorders of the temporomandibular joint may appear, leading to e.g. chronic headaches, tinnitus, and other acoustic symptoms \cite{KIMSJ}. 

The above-described works show that we can automatically search for anatomical landmarks on X-ray cephalometric images with very high accuracy and precision.
Accordingly to \cite{Li}, \cite{Pano-Kim-2}, \cite{Egger}, \cite{Zanjani}, we will be able to quickly classify skeletal or dental defects.
The future of CNNs in X-ray panoramic may be limited due to the increasing displacement of this 2D study by 3D CBCT. This is due to the lower accuracy of 2-dimensional images and the smaller and smaller price difference between individual tests. 
In the case of CBCT examinations, neural networks will help us identify bone pathologies within the maxilla and mandible, determine the bone level, or the location of impacted teeth to qualify for orthodontic treatment. As these are three-dimensional (3D) examinations, we will be able to obtain more accurate results of tooth segmentation from the jaw model. They will significantly reduce the time of manual segmentation, which can be up to 4 hours \cite{Ezhov}.

The greatest challenge is the morphological heterogeneity of humans and the fact that different norms of facial aesthetics are accepted in different cultures. Each patient presents various diseases, takes various medications, is often burdened with genetic disorders, injuries, and hospitalizations, which may affect the therapeutic process. As a result, the number of therapeutic combinations is unlimited \cite{mr}. After collecting the medical history, examining the patient, and performing additional tests, i.e. diagnostic model, CBCT, and cephalometric X-ray, an individual treatment plan is created. Sometimes, a patient may not consent to a proposed treatment plan \cite{RobertsHarry}. The orthodontist may then suggest an alternative treatment plan which further increases the number of therapeutic combinations. Another challenge is the patient himself. The obtained therapeutic effects depend on the cooperation of the patient who, due to insufficient oral hygiene, led to dental caries, periodontal diseases, or, in the case of clear aligner appliances, did not wear them long enough \cite{GkantidisN}. 
For the neural network to be able to effectively solve orthodontic problems, it is necessary to load the appropriate amount of data obtained from patients. In the case of rare abnormalities, the amount of available data will be not sufficient for AI to recognize them - therefore it will take a long time for the neural network to be as efficient at solving clinical problems at the same level of effectiveness as an orthodontist. The last problem in training the neural network is the time the orthodontist has to spend annotating DICOM, JPG, and STL files. 

\section{Discussion}
There are challenges in using datasets annotated by medical professionals. In annotated medical data, many problems can arise which make establishing a gold standard difficult. A dataset of labeled data by two or more medical professionals can have inter-observer variability caused by a difference in years of experience and training. An example of this can be seen in the ISBI 2015 open cephalometric dataset, where images were annotated both by a junior and senior doctor. For the most difficult landmark, inter-observer variability reached 6.57 $\pm$ 0.18 mm \cite{randomforest}. Additionally, for a single person annotating the data, influences such as time constraints can cause intra-observer variability.
Another annotation problem in orthodontics is the time-consuming process of labeling the data. For the segmentation task, many papers cited time-constraints as the reason for a low number of training and validation samples. This is compounded for 3D segmentation data, where there are more pixels to classify \cite{Ezhov}.


During our search, we found little to none multimodal methods which use CNNs in orthodontics. Methods exist which combine a patient's CBCT scan and digital model to improve the reconstruction of a set of teeth, however they do not utilize neural networks in their work \cite {QianJ, ZhouX, YauH}. Other methods combine textual information such as gender and cephalometric x-rays for classification \cite{YuH} or use NLP on doctor’s notes and combine that with images for a diagnosis \cite{Kajiwara}.

\section{Conclusions}

In this paper, we summarize the current application of convolutional neural networks in one of the fields of dentistry - orthodontics - and we show that deep learning methods have a wide application in the orthodontics field. Then, we focus on the main four methods used in orthodontics: cephalometric X-ray imaging, panoramic imaging, cone-beam computed tomography, and dental casts. We give an overview of the various works concerning each method and compare them based on achieved results. The results of the reviewed studies indicate that deep learning methods employed in orthodontics can be far superior in comparison to other high-performing algorithms. Therefore, we believe that deep learning approaches will play a significant role in orthodontics.

\section*{Acknowledgement}
This work was supported by Warsaw University of Technology Dean's grant number II/2019/GD/2 and Grant of Scientific Discipline of Computer Science and Telecommunications at Warsaw University of Technology.



\printcredits

\end{document}